\documentclass[sigconf]{acmart}

\AtBeginDocument{%
  \providecommand\BibTeX{{%
    \normalfont B\kern-0.5em{\scshape i\kern-0.25em b}\kern-0.8em\TeX}}}

\copyrightyear{2020}
\acmYear{2020}
\setcopyright{acmcopyright}\acmConference[KDD '20]{Proceedings of the 26th ACM SIGKDD Conference on Knowledge Discovery and Data Mining USB Stick}{August 23--27, 2020}{Virtual Event, USA}
\acmBooktitle{Proceedings of the 26th ACM SIGKDD Conference on Knowledge Discovery and Data Mining USB Stick (KDD '20), August 23--27, 2020, Virtual Event, USA}
\acmPrice{15.00}
\acmDOI{10.1145/3394486.3403189}
\acmISBN{978-1-4503-7998-4/20/08}

\usepackage{graphicx}
\usepackage{subfigure}
\usepackage{url}
\usepackage{amsmath, amsfonts, amsthm, bm}
\usepackage{soul}
\usepackage{tabularx, multirow}
\usepackage{bm}
\usepackage{dsfont}
\usepackage[ruled,vlined]{algorithm2e}
\usepackage{algpseudocode}

\newcolumntype{P}{>{\centering\arraybackslash}m{0.2\linewidth}}
\newcolumntype{Q}{>{\centering\arraybackslash}m{0.09\linewidth}}
\newcolumntype{R}{>{\centering\arraybackslash}m{0.05\linewidth}}
\newcolumntype{S}{>{\centering\arraybackslash}m{0.25\linewidth}}

\begin{document}
\fancyhead{} 

\title{Multi-Class Data Description for Out-of-distribution Detection}




\author{Dongha Lee, Sehun Yu, Hwanjo Yu}
\authornote{corresponding author}
\affiliation{%
  \institution{Pohang University of Science and Technology (POSTECH), South Korea}
}
\email{{dongha.lee, hunu12, hwanjoyu}@postech.ac.kr}
\newcommand{\smallsection}[1]{{\vspace{0.05in} \noindent \bf {#1.\hspace{5pt}}}}

\newcommand{\mlp}{MLP\xspace}
\newcommand{\cnn}{CNN\xspace}
\newcommand{\relu}{ReLU\xspace}
\newcommand{\resnet}{ResNet\xspace}
\newcommand{\densenet}{DenseNet\xspace}
\newcommand{\adadelta}{AdaDelta\xspace}
\newcommand{\adam}{Adam\xspace}

\newcommand{\proposed}{Deep-MCDD\xspace}
\newcommand{\proposeds}{Deep-MCDDs\xspace}
\newcommand{\dnnlayer}{dm-layer\xspace}
\newcommand{\ocsvm}{OC-SVM\xspace}
\newcommand{\svdd}{SVDD\xspace}
\newcommand{\deepsvdd}{Deep-SVDD\xspace}
\newcommand{\alad}{ALAD\xspace}
\newcommand{\deepncm}{Deep-NCM\xspace}
\newcommand{\protonet}{ProtoNet\xspace}

\newcommand{\ctg}{Cardio\xspace}
\newcommand{\gas}{GasSensor\xspace}
\newcommand{\drive}{DriveDiagnosis\xspace}
\newcommand{\shuttle}{Shuttle\xspace}
\newcommand{\mnist}{MNIST\xspace}
\newcommand{\svhn}{SVHN\xspace}
\newcommand{\cifarten}{CIFAR-10\xspace}
\newcommand{\cifarhundred}{CIFAR-100\xspace}
\newcommand{\tinyimgnet}{TinyImageNet\xspace}
\newcommand{\lsun}{LSUN\xspace}

\newcommand{\sbloss}{\mathcal{L}}
\newcommand{\sdloss}{\mathcal{L}}

\newcommand{\dist}[1]{D_{#1}}
\newcommand{\spradius}[1]{R_{#1}}
\newcommand{\spcenter}[1]{\mathbf{c}_{#1}}
\newcommand{\bias}[1]{b_{#1}}
\newcommand{\mean}[1]{\bm{\mu}_{#1}}
\newcommand{\stdev}[1]{\sigma_{#1}}
\newcommand{\calpha}[2]{\alpha_{#1}^{#2}}
\newcommand{\mapfunc}[1]{f_{#1}}

\newcommand{\tensorflow}{Tensorflow}
\newcommand{\pytorch}{pyTorch}

\begin{abstract}
The capability of reliably detecting out-of-distribution samples is one of the key factors in deploying a good classifier, as the test distribution always does not match with the training distribution in most real-world applications.
In this work, we present a deep multi-class data description, termed as \proposed, which is effective to detect out-of-distribution (OOD) samples as well as classify in-distribution (ID) samples.
Unlike the softmax classifier that only focuses on the linear decision boundary partitioning its latent space into multiple regions, our \proposed aims to find a spherical decision boundary for each class which determines whether a test sample belongs to the class or not.
By integrating the concept of Gaussian discriminant analysis into deep neural networks, we propose a deep learning objective to learn class-conditional distributions that are explicitly modeled as separable Gaussian distributions.
Thereby, we can define the confidence score by the distance of a test sample from each class-conditional distribution, and utilize it for identifying OOD samples.
Our empirical evaluation on multi-class tabular and image datasets demonstrates that \proposed achieves the best performances in distinguishing OOD samples while showing the classification accuracy as high as the other competitors. 

\end{abstract}



\maketitle

\section{Introduction}
\label{sec:intro}

Out-of-distribution (OOD) detection, also known as novelty detection, refers to the task of identifying the samples that differ in some respects from the training samples.
Recently, deep neural networks (DNNs) turned out to show unpredictable behaviors in case of mismatch between the training and test data distributions;
for example, they tend to make high confidence prediction for the samples that are drawn from OOD or that belong to unseen classes~\citep{szegedy2014intriguing, moosavi2017universal}.
For this reason, accurately measuring the \textit{distributional uncertainty}~\citep{malinin2018predictive} of DNNs becomes one of the important challenges in many real-world applications where we can hardly control the test data distribution.
Several recent studies have tried to simply detect OOD samples using the confidence score defined by softmax probability for known classes~\citep{hendrycks2016baseline, liang2017enhancing} or Mahalanobis distance from the class means~\citep{lee2018simple}, and they showed promising results to some extent.

\begin{figure}[t]
	\centering
	\includegraphics[width=\linewidth]{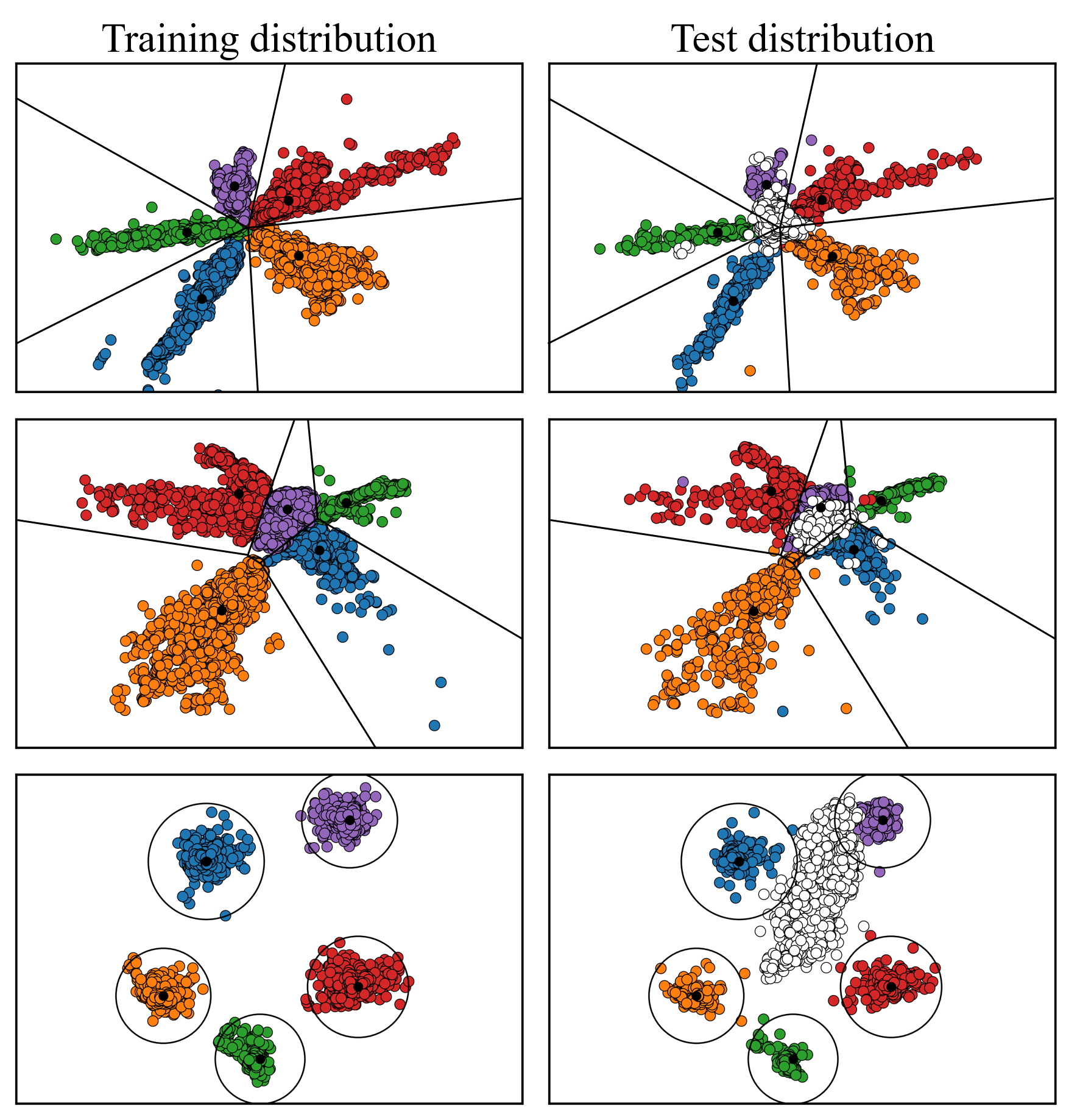}
\caption{The 2D latent spaces and their decision boundaries obtained by DNNs with a softmax classifier (Top), a distance classifier (Middle), and our \proposed (Bottom). In-distribution and out-of-distribution samples are shown as colored and white circles, respectively. Dataset: \gas, OOD class: Ammonia, Model: Multi-layer perceptron.}
\label{fig:latspace}
\end{figure}

However, all of them employ the DNNs trained for a softmax classifier, which has limited power to locate OOD samples distinguishable from in-distribution (ID) samples in their latent space.
To be specific, the softmax classifier is optimized to learn the discriminative latent space where the training samples are aligned along their corresponding class weight vectors, maximizing the softmax probability for the target classes.
As pointed out in~\citep{hendrycks2016baseline}, OOD samples are more likely to have small values of the softmax probability for all known classes, which means that their latent vectors get closer to the origin.
As a result, there could be a large overlap between two sets of ID and OOD samples in the latent space (Figure~\ref{fig:latspace}, Top), which eventually reduces the gap between their confidence scores and degrades the performance as well.

In addition, most of the existing confidence scores adopt additional calibration techniques~\citep{goodfellow2014explaining, hinton2015distilling} to enhance the reliability of the detection, but they include several hyperparameters whose optimal values vary depending on the test data distribution.
In this situation, they utilized a small portion of each test set (containing both ID and OOD samples) for validation, and reported the results evaluated on the rest by using the optimal hyperparameter values for each test case.
However, in practice of OOD detection, prior knowledge of test distributions is not available before we encounter them. 
Thus, such process of tuning the hyperparameters for each test case is not practical when deploying the DNNs in practice.

To overcome the limitations, the goal of our work is to learn the discriminative latent space with the boundaries among ID samples of different classes, and furthermore, with the boundary between ID and OOD samples.
To this end, we propose a deep multi-class data description, termed as \proposed, which is capable of effectively identifying OOD test samples as well as classifying ID test samples.
Inspired by support vector data description (\svdd), a popular kernel method for detecting novel samples based on a spherical boundary, \proposed aims to find multiple spheres as many as the number of given classes so that each of them encloses all training samples of each class (Figure~\ref{fig:latspace}, Bottom).

For sophisticated modeling of the multiple spheres, we represent them as separable class-conditional distributions, assumed to be isotropic Gaussian distributions.
Thus, \proposed places OOD samples further apart from the distributions of all given classes, without utilizing OOD samples for validation.
Based on the distance between a test sample and the obtained class-conditional distributions, we can calculate how likely and how confidently the sample belongs to each class.
Gaussian discriminant analysis (GDA) provides the theoretical background for incorporating the distance classifier into the DNNs.

Our extensive experiments on tabular and image datasets demonstrate that the proposed \proposed classifier identifies OOD samples more accurately than the state-of-the-art methods, while showing the classification accuracy as high as the others.
In addition, qualitative analyses on the latent space obtained by \proposed provide the strong evidence that it places OOD sample more clearly distinguishable from ID samples in the space compared to the existing softmax classifier or distance classifier.

\section{Related Work}
\label{sec:related}

In this section, we survey the literature on most related topics, that are, 1) data description with a spherical boundary, and 2) out-of-distribution (OOD) detection based on deep learning approaches.

\subsection{Hypersphere-based Data Description}
\label{subsec:svdd}
The problem of data description, also known as one-class classification, aims to make a \textit{description} of training data and use it to detect whether a test sample comes from the training distribution or not.
Early work on data description mainly focused on kernel-based approaches~\cite{scholkopfu1999sv, scholkopf2001estimating, tax1999support, tax2004support}, and they tried to represent their decision boundaries by identified support vectors.
Among them, support vector data description (\svdd)~\citep{tax2004support} finds a closed spherical boundary (i.e., hypersphere) that encloses the majority of the data, and the objective is defined by using a center $\mathbf{c}$, a radius $R$, and a feature mapping $\phi$ associated with its kernel.
\begin{equation}
\begin{gathered}
    \min_{R, \mathbf{c}, \xi} R^2 + \frac{1}{\nu N}\sum_{i=1}^{N} \xi_i \\
    \text{s.t.   } \forall i, \text{ }\lVert \phi(x_i) - \mathbf{c}\rVert^2 \leq R^2 + \xi_i, \text{ } \xi_i \geq 0,
\end{gathered}
\end{equation}
where $\xi_i$ is the $i$-th slack variable for a soft boundary and $\nu$ is the regularization coefficient that controls how large the penalty would be allowed, which determines the volume of the sphere.

Recently, \deepsvdd~\cite{ruff2018deep, ruff2019deep} employs deep neural networks (DNNs) instead of kernel methods to effectively capture the normality of high-dimensional data and learn their useful latent representations.
It optimizes the DNNs to map training samples (which are assumed to be from a single known class) close to its center $\mathbf{c}$ in the latent space, showing that it eventually determines a hypersphere of minimum volume with the center $\mathbf{c}$:
\begin{equation}
\label{eq:deepsvdd}
    \min_{\mathcal{W}, \mathbf{c}} \frac{1}{N} \sum_{i=1}^{N} \lVert f(\mathbf{x}_i; \mathcal{W}) - \mathbf{c} \rVert^2.
\end{equation}
However, \deepsvdd is not able to classify input samples into multiple classes given in training data, because it only focuses on detecting novel samples (or outliers), not utilizing class labels at all.

\subsection{Out-of-distribution Detection with DNNs}
As DNNs have become the dominant approach to a wide range of real-world applications and the cost of their errors increases rapidly, many studies have been carried out on measuring the uncertainty of a model's prediction, especially for non-Bayesian DNNs~\cite{gal2016uncertainty, teye2018bayesian}.
Among several types of uncertainty defined in~\cite{malinin2018predictive}, \textit{distributional uncertainty} occurs by the discrepancy between the training and test distributions.
In this sense, the OOD detection task can be understood as modeling the distributional uncertainty, and a variety of approaches have been attempted.
Recently, a generative model-based approach has gained much attention, particularly on the image domain~\cite{salimans2017pixelcnn, kingma2018glow, ren2019likelihood}, because their models can directly compute the likelihood probability of a test sample and use it for OOD detection. 
However, they aim to train the detector which only distinguishes OOD samples from ID samples;
this disables to classify ID samples into given classes.

To enable ID classification and OOD detection with a single model, several recent work started to detect OOD samples by utilizing a trained DNN classifier.
The baseline method~\cite{hendrycks2016baseline} is the first work to define the confidence score by the softmax probability based on a given DNN classifier.
To enhance the reliability of detection, ODIN~\cite{liang2017enhancing} applies two calibration techniques, i.e., temperature scaling~\cite{hinton2015distilling} and input perturbation~\cite{goodfellow2014explaining}, to the baseline method, which can push the softmax scores of ID and OOD samples further apart from each other.
The state-of-the-art method~\cite{lee2018simple} uses the Mahalanobis distance from class means in the latent space, assuming that samples of each class follow the Gaussian distribution with the same covariance in the latent space.

However, the existing OOD detection methods using a DNN classifier still have several limitations.
First, all of them utilize the DNNs trained for the softmax classifier (i.e., only optimized for classifying ID samples), so they have difficulty in distinguishing OOD samples from ID samples in their latent space (Figure~\ref{fig:latspace}, Top). 
It is worth noting that using the Mahalanobis distance based on the same covariance matrix for all classes~\cite{lee2018simple} also hinders to accurately measure the confidence score, since the DNNs are not trained to satisfy such condition of the same covariance matrix.
In addition, the calibration techniques for their confidence score include several hyperparameters which need to be tuned by using the OOD samples from the test distribution;
this makes the detector less practical, because test distributions are assumed to be unknown in many real-world applications.

\section{Deep Multi-class Data Description}
\label{sec:method}
In this section, we introduce \proposed, a multi-class data description based on DNNs, to accurately detect out-of-distribution samples as well as classify in-distribution samples into known classes.
We first present the learning objective of \proposed and its theoretical backgrounds in the perspective of Gaussian discriminant analysis (GDA), then propose the distance-based classifier and OOD detector.

\begin{figure}[t]
	\centering
	\includegraphics[width=\linewidth]{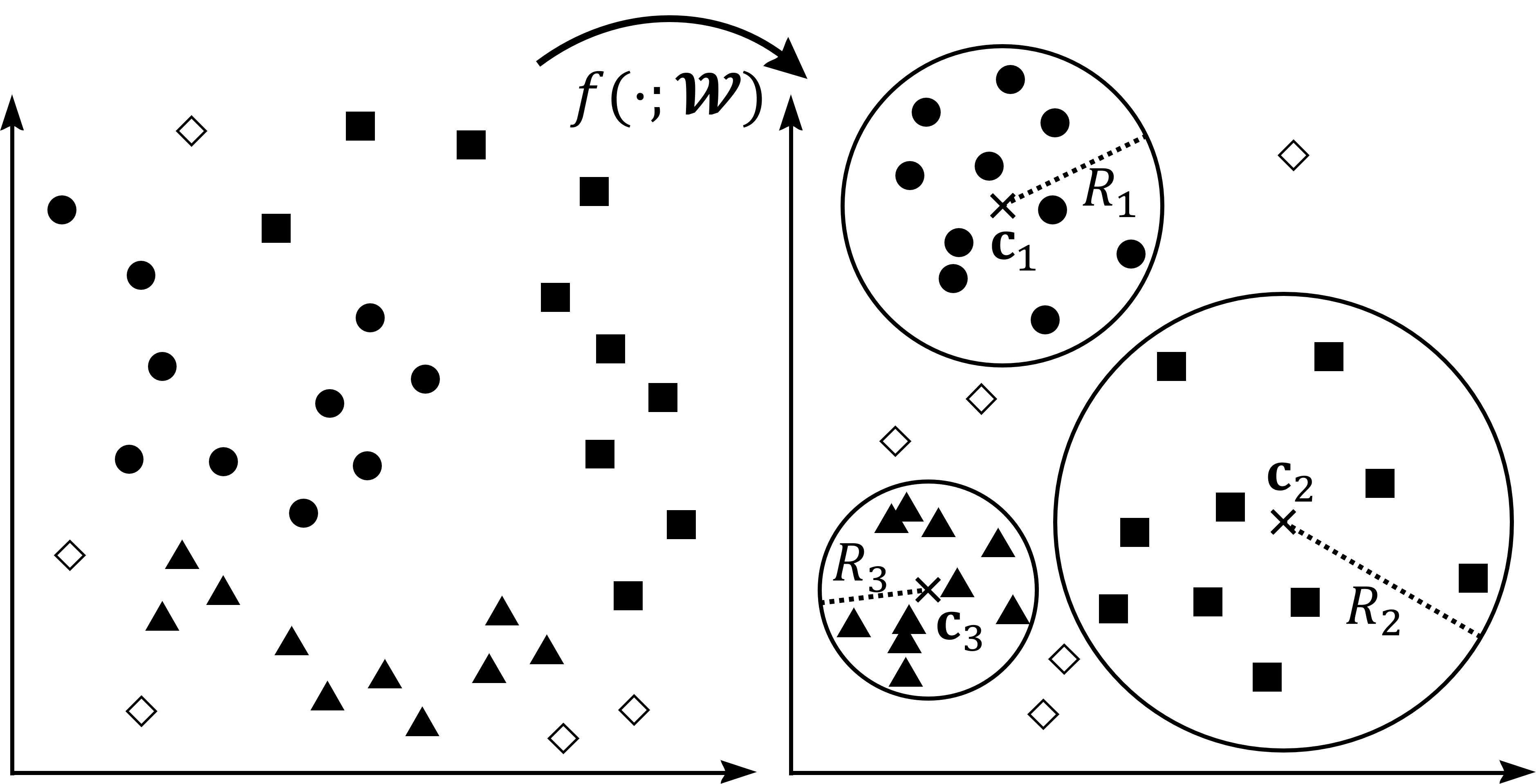}
\caption{The overview of deep multi-class data description. \proposed induces the latent space where ID samples (represented by black markers) are inside the hyperspheres corresponding to their classes and OOD samples (represented by white diamonds) fall outside all the hyperspheres.}
\label{fig:overview}
\end{figure}

\subsection{The \proposed Objective}
\label{subsec:obj}
\subsubsection{Modeling class boundaries as multiple hyperspheres}
The key idea of our objective is to optimize DNNs so that the latent representations (i.e., the outputs of the last layer) of data samples in the same class gather together thereby form an independent sphere of minimum volume (Figure \ref{fig:overview}).
In other words, \proposed incorporates $K$ one-class classifiers modeled by hyperspheres into a single network.
The purpose of learning a single hypersphere is to determine whether a test sample belong to the target class or not, thus training it for each class is useful for detecting OOD samples that do not belong to any given classes.

Let $f(\cdot;\mathcal{W})$ be a DNN with the set of network parameters $\mathcal{W}$, and the spherical boundary for class $k$ be specified by its center $\mathbf{c}_k$ and radius $R_k$.
Given $N$ training samples $\{(\mathbf{x}_1, y_1), \ldots, (\mathbf{x}_N, y_N)\}$ from $K$ different classes, the objective of \textit{soft-boundary} \proposed is described as follows.
\begin{equation}
\label{eq:softmcdd}
    \min_{\mathcal{W}, \textbf{c}, R} \sum_{k=1}^{K} \left[ \spradius{k}^2 + \frac{1}{\nu N}\sum_{i=1}^N \max \left\{0, \alpha_{ik} \left(\lVert  f(\mathbf{x}_i; \mathcal{W}) - \spcenter{k} \rVert^2 - \spradius{k}^2 \right) \right\} \right],
\end{equation}
where $\alpha_{ik}$ denotes the class assignment indicator whose value is $1$ if $\mathbf{x}_i$ belongs to class $k$, and $-1$ otherwise.
As done in~\cite{ruff2018deep}, this objective includes the penalty term for the samples that fall outside its class decision boundary (i.e., the samples that violate $\lVert f(\mathbf{x}_i; \mathcal{W}) -\spcenter{y_i} \rVert < R_{y_i}$), but it additionally considers the distances between samples and all the other class centers for multi-class classification.
That is, new constraints $\lVert f(\mathbf{x}_i; \mathcal{W}) - \spcenter{k} \rVert> R_{k}, \forall k\neq y_i$ are added so that each sample is located outside the spheres that the sample does not belong to.
Note that $\nu$ determines how strongly each hypersphere encloses the corresponding training samples, which is eventually related to the violations of the class boundary and the volume of the sphere.

However, the optimization of \eqref{eq:softmcdd} by stochastic gradient descent (SGD) is challenging due to the different scales of $\mathcal{W}$, $\mathbf{c}$, and $R$.
For this reason, we use the block coordinate descent (BCD) approach, which alternately updates the set of target parameters to minimize the objective while fixing the others.
In detail, we train the network parameters while the class centers and radii are fixed, then all the class centers and radii are updated based on the fixed DNN.

\subsubsection{Modeling discriminative class-conditional distributions}
To facilitate the optimization and model the multiple spheres in a more sophisticated manner, we introduce an alternative objective to learn multiple Gaussian distributions instead of the hyperspheres.
Let samples of class $k$ be generated from a $k$-th class-conditional distribution, which is entirely separable from other class-conditional distributions.
We consider each of them as an isotropic Gaussian distribution with class mean $\mean{k}$ and standard deviation $\stdev{k}$ in the latent space, denoted by $\mathcal{N}(\mean{k}, \stdev{k}^2I)$.
Using the obtained distributions, we can calculate the class-conditional probabilities that indicate how likely an input sample is generated from each distribution, and this probability can serve as a good measure of the confidence.
We first define the distance function $\dist{k}(\cdot)$ based on the $k$-th class-conditional distribution:
\begin{equation}
\begin{split}
    \dist{k}(\mathbf{x}) &= - \log P(\mathbf{x} | y=k) \\
    &= - \log \mathcal{N}(f(\mathbf{x}; \mathcal{W}) \mid \mean{k}, \stdev{k}^2 I) \\
    &\approx \frac{\lVert f(\mathbf{x}; \mathcal{W}) - \mean{k} \rVert^2}{2\stdev{k}^2} + \log \stdev{k}^d
\end{split}
\label{eq:distance}
\end{equation}
We cancel out the class-independent constant $\log (2\pi)^{d/2}$ and add a non-negative constraint for log standard deviations, $\log \stdev{k} \geq 0$, in order to make the distance function satisfy triangular inequality, which is one of the conditions to be a \textit{metric}.
Note that $\spcenter{k}$ and $\spradius{k}$ used for modeling the hypersphere in \eqref{eq:softmcdd} are respectively replaced with $\mean{k}$ and $\stdev{k}$, and the standard deviations are further taken into account for computing the distances.

Using the class-specific distance function, we propose the \proposed objective which 1) matches the empirical distributions of training samples with the target class-conditional distributions in the latent space, and simultaneously 2) makes them distinguishable from each other for multi-class classification.
The objective is described as
\begin{equation}
    \min_{\mathcal{W}, \bm{\mu}, \sigma, b} \frac{1}{N} \sum_{i=1}^N  \left[ \dist{y_i}(\mathbf{x}_i) -\frac{1}{\nu} \log \frac{\exp \left( - \dist{y_i}(\mathbf{x}_i) + \bias{y_i} \right)}{\sum_{k=1}^{K} \exp \left( - \dist{k}(\mathbf{x}_i) + \bias{k}\right)} \right] .
\label{eq:obj}
\end{equation}
As defined in \eqref{eq:softmcdd}, the regularization coefficient $\nu$ controls the importance of the two terms;
each of them affects the variance of the Gaussian distributions and their classification power, respectively.
The objective includes four types of trainable parameters: the weights of a DNN $\mathcal{W}$, the class means $\mean{1},\ldots,\mean{K}$, standard deviations $\stdev{1},\ldots,\stdev{K}$, and biases $b_1,\ldots,b_K$.
All of them can be effectively optimized at the same time by using minibatch SGD and gradient back-propagation, which are conventionally used for training DNNs.
Unlike other DNN classifiers that need to explicitly compute the class means by their definitions $\mean{k} = \frac{1}{N_k}\sum_{y_i=k} f(\mathbf{x}_i)$ where $N_k$ is the number of training samples of class $k$~\cite{snell2017prototypical, guerriero2018deep}, \proposed obtains approximated ones without further computations while optimizing the DNN.


\subsection{Theoretical Backgrounds}
\label{subsec:theory}
The \proposed objective \eqref{eq:obj} can be understood from the perspective of Gaussian discriminant analysis (GDA).
The conventional discriminative (or softmax) classifier widely used for DNNs is optimized to maximize the softmax probability for the target class, computed by its fully-connected layer.  
Unlike the discriminative classifier, the generative classifier defines the posterior distribution $P(y|\mathbf{x})$ by using the class-conditional distribution $P(\mathbf{x}|y)$ and class prior $P(y)$.
In case of GDA, each class-conditional distribution is assumed to follow the multivariate Gaussian distribution\footnote{As the goal of our work is to employ DNNs in the GDA framework, we consider the class-conditional distributions of the latent representations produced by the DNNs.} (i.e., $P(\mathbf{x}|y=k)=\mathcal{N}(f(\mathbf{x})|\mean{k}, \Sigma_k)$) and the class prior is assumed to follow the Bernoulli distribution (i.e., $P(y=k)=\frac{\beta_k}{\sum_{k'}\beta_{k'}}$).
To simply fuse GDA with DNNs, we further assume each class covariance to be isotropic with its standard deviation (i.e., $\Sigma_k=\stdev{k}^2 I$).
Then, the posterior probability that a sample $\mathbf{x}$ belongs to the class $k$ is described as 
\begin{equation*}
\begin{split}
    P(y=k|\mathbf{x}) &= \frac{P(y=k) P(\mathbf{x}|y=k)}{\sum_{k'} P(y=k') P(\mathbf{x} | y=k')} \\
    &= \frac{ \exp\left(-(2\stdev{k}^2)^{-1} \lVert f(\mathbf{x})- \mean{k} \rVert^2 -\log \stdev{k}^d + \log \beta_k\right)}{\sum_{k'} \exp\left(-(2\stdev{k'}^2)^{-1} \lVert f(\mathbf{x}) -\mean{k'}\rVert^2 -\log \stdev{k'}^d + \log \beta_{k'}\right)}.
\end{split}
\label{eq:gda}
\end{equation*}
Considering $\log\beta_{k}$ as the class bias $b_{k}$, the second term of our objective~\eqref{eq:obj} is equivalent to the negative log posterior probability.
In other words, the objective trains the classifier so that it maximizes the posterior probability of training samples for their target classes.

However, the direct optimization of the DNNs and other parameters by SGD does not guarantee that the class-conditional distributions become the Gaussian distributions and the obtained class means match with the actual class means of training samples.
Thus, to enforce our GDA assumption, we minimize the Kullback-Leibler (KL) divergence between the $k$-th empirical class-conditional distribution and the Gaussian distribution whose mean and covariance are $\mean{k}$ and $\stdev{k}^2 I$, respectively.
The empirical class-conditional distribution is represented by the average of the dirac delta functions for all training samples of a target class, i.e., $\mathbb{P}_k = \frac{1}{N_k} \sum_{y_i = k}\delta(\mathbf{x} - f(\mathbf{x}_i))$.
Then, the KL divergence is formulated as
\begin{equation*}
\begin{split}
    &\mathrm{KL}(\mathbb{P}_{k} \parallel \mathcal{N}(\mean{k}, \stdev{k}^2 I)) \\
    &= - \int \frac{1}{N_k} \sum_{y_i = k}\delta(\mathbf{x} - f(\mathbf{x}_i)) \log \left[ \frac{1}{(2\pi\stdev{k}^2)^{d/2}}\exp\left(-\frac{\lVert \mathbf{x} - \mean{k} \rVert^2}{2\stdev{k}^2} \right)\right] \mathrm{d}\mathbf{x} \\
    &\quad + \int \frac{1}{N_k} \sum_{y_i = k}\delta(\mathbf{x} - f(\mathbf{x}_i)) \log \left[ \frac{1}{N_k} \sum_{y_i = k}\delta(\mathbf{x} - f(\mathbf{x}_i)) \right] \mathrm{d}\mathbf{x}\\
    &= -\frac{1}{N_k}\sum_{y_i = k} \log \left[ \frac{1}{(2\pi\stdev{k}^2)^{d/2}}\exp\left(-\frac{\lVert f(\mathbf{x}_i) -\mean{k} \rVert^2}{2\stdev{k}^2} \right)\right] + \log \frac{1}{N_k} \\
    & = \frac{1}{N_k} \sum_{y_i = k} \left( \frac{\lVert f(\mathbf{x}_i) - \mean{k}\rVert^2}{2\stdev{k}^2} + \log \stdev{k}^d \right) + \mathrm{constant}.
\end{split}
\label{eq:klreg}
\end{equation*}
The entropy term of the empirical class-conditional distribution can be calculated by using the definition of the dirac measure.
By minimizing this KL divergence for all the classes, we can approximate the $K$ class-conditional Gaussian distributions.
Finally, we complete our objective by combining this KL term with the posterior term using the $\nu$-weighted sum in order to control the effect of the regularization.
We remark that $\nu$ is the hyperparameter used for training the model, which depends on only ID, not OOD;
thus it does not need to be tuned for different test distributions.

\subsection{ID Classification and OOD Detection}
\subsubsection{In-distribution classification}
Since our objective maximizes the posterior probability for the target class of each training sample $P(y=y_i|\mathbf{x})$, we can predict the class label of an input sample to the class that has the highest posterior probability as follows.
\begin{equation}
\begin{split}
    \hat{y}(\mathbf{x}) &= \arg\max_k P(y=k|\mathbf{x}) \\
    &= \arg\max_k \left[- \dist{k}(\mathbf{x}) +\bias{k} \right]
\end{split}
\label{eq:classification}
\end{equation}
In terms of DNNs, our proposed classifier replaces the fully-connected layer (fc-layer) computing the final classification score by $\mathbf{w}_k \cdot f(\mathbf{x})+b_k$ with the \textit{distance metric layer} (dm-layer) computing the distance from each class mean by $-\dist{k}(\mathbf{x}) + b_k$.
In other words, the class label is mainly predicted by the distance from each distribution, so our classifier can be categorized as ``distance classifier''. 

\subsubsection{Out-of-distribution detection}
Using the trained \proposed, the confidence score of each sample can be computed based on the class-conditional probability $P(\mathbf{x}|y=k)$.
Taking the log of the probability, we simply define the confidence score $S(\mathbf{x})$ using the distance between a test sample and the closest class-conditional distribution in the latent space,
\begin{equation}
\label{eq:detection}
    S(\mathbf{x}) = - \min_k \dist{k}(\mathbf{x}).
\end{equation}
This distance-based confidence score yields discriminative values between ID and OOD samples.
In the experiment section, we show that the proposed distance in the latent space of \proposed is more effective to detect the samples not belonging to the $K$ classes, compared to the Mahalanobis distance in the latent space of the softmax classifier.
Moreover, it does not require additional computation to obtain the class means and covariance matrix, and the uncertainty can be measured by a single DNN inference.


\section{Experiments}
\label{sec:exp}

In this section, we present experimental results that support the superiority of \proposed.
Using tabular and image datasets, we mainly compare the performance of our distance classifier (i.e., DNNs with dm-layer) with that of the softmax classifier (i.e., DNNs with fc-layer) in terms of both ID classification and OOD detection.
We also validate the effectiveness of \proposed against existing distance classifiers as well as OOD detectors, then provide empirical analyses on the effect of the regularization term.

\begin{table}[t]
    \caption{Statistics of tabular datasets.}
    \label{tbl:dataset}
    \centering
    \begin{tabular}{cPPP}
    \hline
    Dataset & \# Attributes & \# Instances & \# Classes \\\hline
    \gas & 128 & 13,910 & 6 \\
    \shuttle & 9 & 58,000 & 7 \\
    \drive & 48 & 58,509 & 11 \\
    \mnist & 784 & 70,000 & 10 \\\hline
    \end{tabular}
\end{table}

\subsection{Evaluation on Tabular Datasets}
\label{subsec:tbleval}
\smallsection{Experimental settings}
We first evaluate our \proposed using four multi-class tabular datasets with real-valued attributes: \gas, \shuttle, \drive, and \mnist.
They are downloaded from UCI Machine Learning repository\footnote{https://archive.ics.uci.edu/ml/index.php}, and we use them after preprocessing all the attributes based on the z-score normalization.
Table~\ref{tbl:dataset} summarizes the details of the datasets.
To simulate the scenario that the test distribution includes both ID and OOD samples, we build the training and test set by regarding one of classes as the OOD class and the rest of them as the ID classes.
We exclude the samples of the OOD class from the training set, then train the DNNs using only the ID samples for classifying inputs into the $K$-1 classes.
The test set contains all samples of the OOD class as well as the ID samples that are left out for testing.
The evaluations are repeated while alternately changing the OOD class, thus we consider $K$ scenarios for each dataset.
For all the scenarios, we perform 5-fold cross validation and report the average results.

\begin{table*}[t]
    \caption{Performance of ID classification and OOD detection on tabular datasets. The best results are marked in bold face.}
    \label{tbl:tblresults}
    \centering
        \begin{tabular}{cRccccc}
        \hline
        \multirow{2}{*}{Dataset} & \multirow{2}{*}{OOD} & Classification acc. & TNR at TPR 85\% & AUROC & AUPR & Detection acc. \\\cline{3-7}
        & & \multicolumn{5}{c}{Softmax~\cite{hendrycks2016baseline} / Mahalanobis~\cite{lee2018simple} / \proposed}  \\\hline
        
        \multirow{6}{*}{\rotatebox{90}{\gas}} 
        & 0 & 99.65 / 99.35 / 99.54 & 42.42 / \textbf{95.95} / 94.07 & 57.46 / 95.90 / \textbf{96.08} &  46.47 / \textbf{96.65} / 95.70 & 64.99 / \textbf{91.52} / 90.85 \\
        & 1 & 99.71 / 99.33 / 99.57 & 35.00 / 87.82 / \textbf{99.46} & 49.54 / 94.86 / \textbf{99.04} &  39.38 / 93.76 / \textbf{98.97} & 60.96 / 88.65 / \textbf{96.45} \\
        & 2 & 99.74 / 99.53 / 99.66 & 72.63 / 88.97 / \textbf{95.70} & 85.48 / 92.97 / \textbf{97.01} &  83.59 / 95.31 / \textbf{97.58} & 79.42 / 87.94 / \textbf{92.24} \\
        & 3 & 99.67 / 99.42 / 99.56 & 93.97 / 33.80 / \textbf{99.99} & 95.87 / 75.88 / \textbf{99.65} &  95.28 / 84.13 / \textbf{99.78} & 90.36 / 72.73 / \textbf{99.05} \\
        & 4 & 99.72 / 99.54 / 99.66 & 67.31 / 94.99 / \textbf{99.55} & 78.67 / 96.65 / \textbf{99.15} &  59.75 / 96.42 / \textbf{99.08} & 76.62 / 90.94 / \textbf{96.60} \\
        & 5 & 99.70 / 99.41 / 99.58 & 47.44 / 19.05 / \textbf{96.57} & 78.72 / 69.88 / \textbf{97.04} &  81.00 / 79.53 / \textbf{97.83} & 74.50 / 69.73 / \textbf{92.54} \\\hline
        
        \multirow{7}{*}{\rotatebox{90}{\shuttle}} 
        & 0 & 99.94 / 99.94 / 99.89 & 88.69 / \textbf{99.36} / 99.09 & 91.69 / \textbf{99.11} / \textbf{99.12} & 25.85 / \textbf{93.84} / 91.45 & 90.27 / \textbf{98.10} / 97.85 \\
        & 1 & 99.96  / 99.93 / 99.91 & 69.20 / \textbf{100.0} / \textbf{100.0} & 77.34 / 99.58 / \textbf{99.76} & 99.66 / \textbf{99.99} / \textbf{99.99} & 79.94 / 99.22 / \textbf{99.47} \\
        & 2 & 99.96 / 99.96 / 99.93 & 52.63 / \textbf{98.83} / 95.67 & 63.63 / \textbf{98.51} / 97.54 & 97.68 / \textbf{99.96} / \textbf{99.95} & 70.58 / \textbf{94.75} / 92.96 \\
        & 3 & 99.96 / 99.93 / 99.90 & 96.42 / 98.21 / \textbf{99.95} & 98.08 / 98.41 / \textbf{99.77} & 98.07 / 98.67 / \textbf{99.82} & 92.68 / 94.61 / \textbf{98.69} \\
        & 4 & 99.97 / 99.96 / 99.93 & 69.80 / \textbf{100.0} / \textbf{100.0}  & 76.66 / \textbf{99.90} / \textbf{99.91} & 85.19 / \textbf{99.97} / \textbf{99.97} & 80.81 / \textbf{99.87} / \textbf{99.75} \\
        & 5 & 99.95 / 99.93 / 99.92 & 14.00 / \textbf{100.0} / \textbf{100.0}  & 16.84 / \textbf{99.94} / \textbf{99.98} & 99.49 / \textbf{100.0} / \textbf{100.0} & 56.01 / \textbf{99.92} / \textbf{99.97} \\
        & 6 & 99.97 / 99.93 / 99.93 & 00.00 / 96.92 / \textbf{100.0} & 00.00 / 96.82 / \textbf{99.93} & 99.24 / 99.97 / \textbf{100.0} & 50.00 / 98.24 / \textbf{99.96} \\\hline
        
        \multirow{11}{*}{\rotatebox{90}{\drive}}
        & 0 & 99.71 / 98.47 / 99.67 & 13.74 / \textbf{82.22} / 69.23 & 20.78 / \textbf{91.28} / 83.65 & 50.16 / \textbf{95.97} / 88.71 & 51.70 / \textbf{84.63} / 77.59 \\
        & 1 & 99.72 / 98.89 / 99.74 & 07.98 / \textbf{55.94} / 50.86 & 12.09 / \textbf{79.74} / 78.15 & 47.88 / \textbf{88.16} / 87.42 & 50.04 / \textbf{74.44} / 71.72 \\
        & 2 & 99.67 / 98.29 / 99.65 & 63.73 / 55.50 / \textbf{88.93} & 75.30 / 79.92 / \textbf{93.98} & 77.63 / 88.96 / \textbf{96.39} & 75.96 / 73.65 / \textbf{87.53} \\
        & 3 & 99.67 / 98.20 / 99.58 & 53.78 / 80.39 / \textbf{94.47} & 63.16 / 90.43 / \textbf{96.42} & 67.22 / 95.63 / \textbf{98.00} & 69.68 / 84.57 / \textbf{91.23} \\
        & 4 & 99.74 / 98.59 / 99.67 & 78.71 / 23.42 / \textbf{88.58} & 81.72 / 66.72 / \textbf{94.04} & 78.87 / 81.82 / \textbf{96.46} & 82.44 / 64.03 / \textbf{87.78} \\
        & 5 & 99.75 / 98.67 / 99.71 & 68.82 / 24.51 / \textbf{76.32} & 78.24 / 68.24 / \textbf{87.06} & 77.24 / 83.20 / \textbf{90.14} & 77.58 / 65.46 / \textbf{80.92} \\
        & 6 & 99.63 / 98.36 / 99.56 & 08.63 / \textbf{99.83} / 98.59 & 10.58 / \textbf{99.67} / 98.59 & 47.49 / \textbf{99.86} / 99.27 & 50.92 / \textbf{98.16} / 93.98 \\
        & 7 & 99.68 / 98.31 / 99.64 & 24.62 / \textbf{66.13} / 44.31 & 34.04 / \textbf{85.34} / 72.38 & 54.50 / \textbf{93.35} / 82.89 & 55.47 / \textbf{79.10} / 68.90 \\
        & 8 & 99.68 / 98.57 / 99.65 & 59.24 / 43.86 / \textbf{79.19} & 74.69 / 75.40 / \textbf{89.88} & 76.65 / 86.17 / \textbf{94.15} & 74.10 / 69.62 / \textbf{82.38} \\
        & 9 & 99.70 / 98.94 / 99.70 & 02.38 / \textbf{65.51} / 28.35 & 04.43 / \textbf{84.74} / 62.83 & 46.05 / \textbf{91.87} / 76.49 & 50.00 / \textbf{77.47} / 61.48 \\
        & 10 & 99.61 / 98.23 / 99.58 & 10.06 / \textbf{100.0} / 99.76 & 12.93 / \textbf{99.97} / 99.57 & 48.24 / \textbf{99.99} / \textbf{99.78} & 51.89 / \textbf{99.97} / 97.26 \\\hline
        
        \multirow{10}{*}{\rotatebox{90}{\mnist}}
        & 0 & 97.82 / 96.93 / 96.58 & \textbf{85.81} / 63.55 / 85.23 & 82.93 / 84.10 / \textbf{91.03} & 78.65 / 91.25 / \textbf{93.37} & \textbf{87.34} / 76.97 / 85.40 \\
        & 1 & 97.86 / 96.87 / 96.45 & \textbf{93.34} / 09.77 / 79.04 & \textbf{90.39} / 59.52 / 88.26 & 85.29 / 75.18 / \textbf{91.20} & \textbf{91.47} / 61.69 / 82.63 \\
        & 2 & 97.97 / 97.16 / 96.84 & 73.88 / 74.99 / \textbf{84.47} & 71.61 / 88.83 / \textbf{89.19} & 69.30 / 89.98 / \textbf{91.70} & 82.52 / 81.38 / \textbf{84.81} \\
        & 3 & 97.99 / 97.50 / 96.87 & 72.31 / 49.71 / \textbf{81.24} & 69.82 / 79.19 / \textbf{86.33} & 67.77 / 88.12 / \textbf{89.75} & 81.29 / 72.72 / \textbf{83.23} \\
        & 4 & 98.02 / 97.26 / 95.02 & 51.20 / 24.91 / \textbf{63.12} & 49.36 / 65.96 / \textbf{81.19} & 58.26 / 80.39 / \textbf{86.15} & 71.27 / 62.61 / \textbf{74.96} \\
        & 5 & 98.04 / 97.38 / 96.98 & 74.38 / 37.44 / \textbf{88.09} & 72.18 / 74.37 / \textbf{92.22} & 71.58 / 86.99 / \textbf{95.30} & 82.98 / 69.30 / \textbf{86.65} \\
        & 6 & 97.86 / 96.97 / 96.47 & 73.86 / 49.43 / \textbf{79.96} & 71.30 / 79.04 / \textbf{84.75} & 69.57 / 88.40 / \textbf{87.92} & 81.82 / 72.26 / \textbf{82.90} \\
        & 7 & 98.10 / 97.17 / 96.97 & 71.20 / 46.75 / \textbf{81.82} & 68.90 / 77.26 / \textbf{89.96} & 66.75 / 86.20 / \textbf{92.75} & 81.10 / 70.32 / \textbf{83.52} \\
        & 8 & 98.10 / 97.41 / 97.28 & 86.21 / 16.74 / \textbf{94.75} & 83.78 / 62.85 / \textbf{95.29} & 79.79 / 79.81 / \textbf{97.34} & 87.63 / 63.00 / \textbf{90.57} \\
        & 9 & 98.21 / 97.42 / 96.99 & 78.06 / 12.69 / \textbf{79.93} & 75.95 / 57.82 / \textbf{88.63} & 73.38 / 76.29 / \textbf{92.14} & 80.44 / 60.40 / \textbf{82.75} \\\hline
        \end{tabular}
\end{table*}

\smallsection{Implementation details}
The multi-layer perceptron (MLP) with three hidden layers is chosen as the DNNs for training the tabular data.
For fair comparisons, we employ the same architecture of MLP (\# Input attributes $\times 128 \times 128 \times 128 \times$ \# Classes) for both the softmax classifier and the distance classifier.
We use the \adam optimizer~\citep{kingma2014adam} with the initial learning rate $\eta=0.01$, and set the maximum number of epochs to $100$.
In case of tabular data, we empirically found that the regularization coefficient $\nu$ hardly affects the performance of our model, so fix it to $1.0$ without further hyperparameter tuning.
To enforce non-negativity for the log standard deviation of each class-conditional distribution, we simply use the ReLU activation.

\smallsection{Evaluation metrics}
For evaluation, we measure the classification accuracy for ID test samples,
as well as four performance metrics for OOD detection: the true negative rate (TNR) at 85\% true positive rate (TPR), the area under the receiver operating characteristic curve (AUROC), the area under the precision-recall curve (AUPR), and the detection accuracy.\footnote{These performance metrics have been mainly used for OOD detection~\citep{lee2018simple, liang2017enhancing}. We measure the AUPR values where ID samples are specified as positive.}
They are mainly used to evaluate the effectiveness of the confidence scores in distinguishing ID and OOD samples.

\subsubsection{Comparison with classifier-based OOD detectors}
We consider two competing methods using the DNNs optimized for the softmax classifier: 1) the baseline method \citep{hendrycks2016baseline} uses a maximum value of softmax posterior probability as the confidence score, $\max_k \frac{\exp(\mathbf{w}_k^\top \cdot f(\mathbf{x}) + b_k)}{\sum_{k'}\exp(\mathbf{w}_{k'}^\top \cdot f(\mathbf{x}) + b_{k'})}$, and 2) the state-of-the-art method~\citep{lee2018simple} defines the score based on the Mahalanobis distance from the closest class mean, $-\min_k (f(\mathbf{x}) - \hat{\mu}_k)^\top \hat{\Sigma}^{-1} (f(\mathbf{x}) - \hat{\mu}_k)$, where $\hat{\mu}_k$ is an empirical class mean and $\hat{\Sigma}$ is a tied-covariance matrix.\footnote{The state-of-the-art method also can predict the class label of test samples by the Mahalanobis distance from class means, $\hat{y}(\mathbf{x})=\arg\min_k (f(\mathbf{x}) - \hat{\mu}_k)^\top \hat{\Sigma}^{-1} (f(\mathbf{x}) - \hat{\mu}_k)$.}
Note that any OOD samples are not available at training time, so we do not consider advanced calibration techniques for all the methods; for example, temperature scaling, input perturbation~\citep{liang2017enhancing, devries2018learning}, and regression-based feature ensemble~\citep{lee2018simple}.

In Table~\ref{tbl:tblresults}, \proposed that defines the classification/confidence scores based on the proposed distance function~\eqref{eq:distance} considerably outperforms the other competing methods using the softmax classifier in most scenarios.
Compared to the baseline method, the Mahalanobis distance-based confidence score sometimes performs better, and sometimes worse.
This strongly indicates that the empirical data distribution in the latent space does not always take the form of a Gaussian distribution for each class, in case of the softmax classifier.
For this reason, our explicit modeling of class-conditional Gaussian distributions using the dm-layer guarantees the GDA assumption, and it eventually helps to distinguish OOD samples from ID samples.
Moreover, our distance classifier shows almost the same classification accuracy with the softmax classifier;
that is, it improves the performance of OOD detection without compromising the performance of ID classification.

For qualitative comparison on the latent spaces of the softmax classifier and distance classifier, we plot the 2D latent space after training the DNNs whose size of latent dimension is set to 2.
Figure~\ref{fig:latspace} illustrates the training and test distributions of the \gas dataset, where the class 3 (i.e., Ammonia) is considered as the OOD class.
Our DNNs successfully learn the latent space so that ID and OOD samples are separated more clearly than the DNNs of the softmax classifier.
Notably, in case of the softmax classifier, the covariance matrices of all the classes are not identical, which violates the necessary condition for the Mahalanobis distance-based confidence score to be effective in detecting OOD samples.\footnote{This confidence score is derived under the assumption that all the classes share the same covariance matrix.}
In this sense, \proposed does not require such assumption any longer, because our objective makes the latent space satisfy the GDA assumption.

\subsubsection{Comparison with non-classifier-based OOD detectors}
\label{subsubsec:compdetector}
We compare the performances of the proposed methods, including the soft-boundary \proposed from Equation~\eqref{eq:softmcdd} (denoted by Soft-MCDD), with those of other OOD detection methods that only learns the boundary between ID and OOD samples.
Among the methods, we choose \deepsvdd~\cite{ruff2018deep} and \alad~\cite{zenati2018adversarially} as our baselines; 
they focus on learning the normality of ID samples without using their class labels.
For the confidence score, \deepsvdd uses the Euclidean distance from the center of a unified hypersphere in the latent space, and \alad measures the magnitude of error between an input sample and the reconstructed one obtained by generative adversarial networks (GANs). 
Note that these \textit{non-classifier}-based OOD detectors are not able to tell each class from the others.
In this experiment, we report the averaged results over $K$ number of OOD scenarios for each tabular dataset.

\begin{figure*}[t]
	\centering
	\begin{minipage}{0.48\linewidth}
	    \centering
        \includegraphics[width=\linewidth]{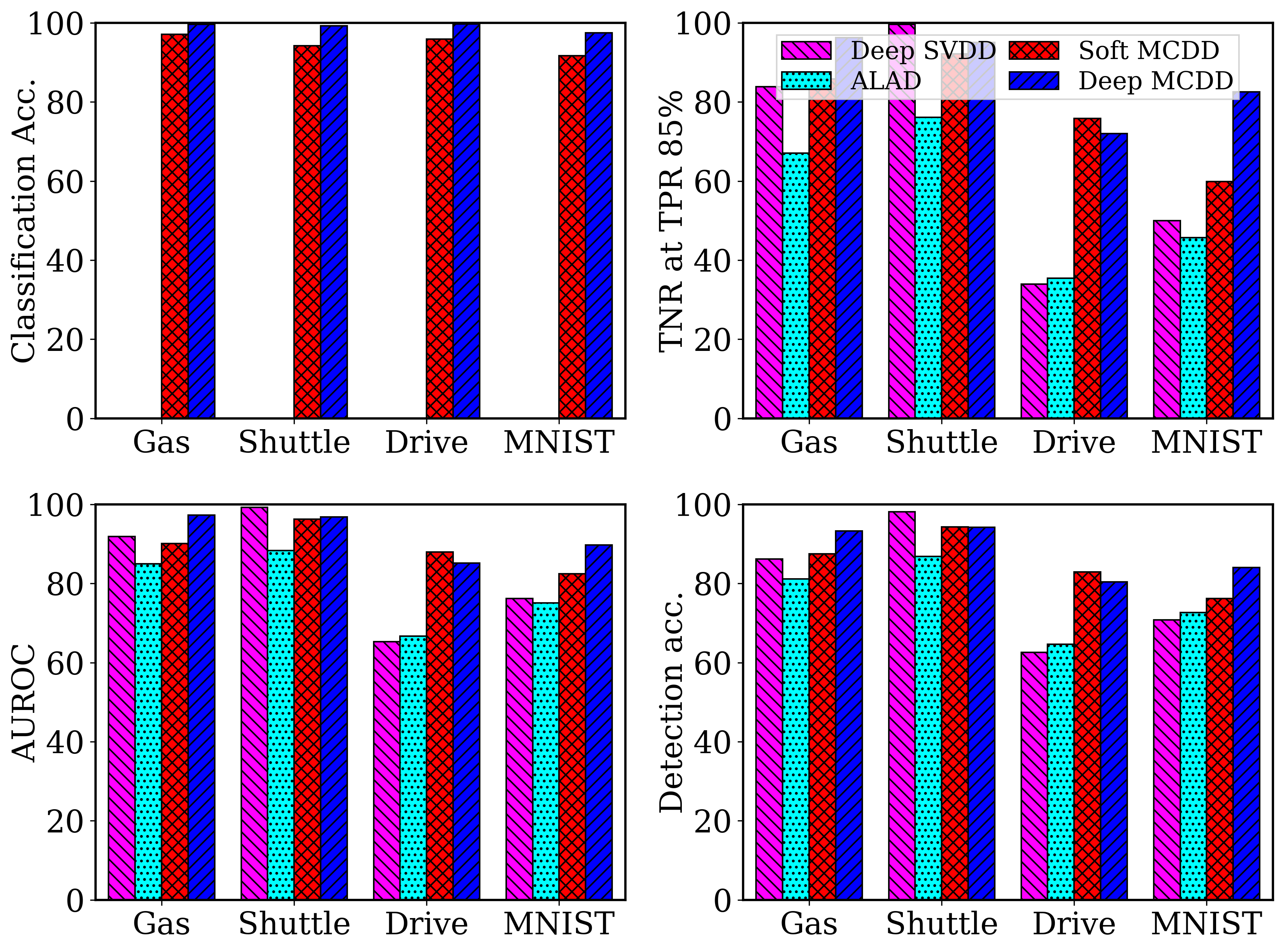}
        \caption{Comparison with non-classifier-based OOD detectors which do not utilize any class information for training.}
        \label{fig:compdetector}
	\end{minipage}
	\hspace{10pt}
	\begin{minipage}{0.48\linewidth}
        \centering
        \includegraphics[width=\linewidth]{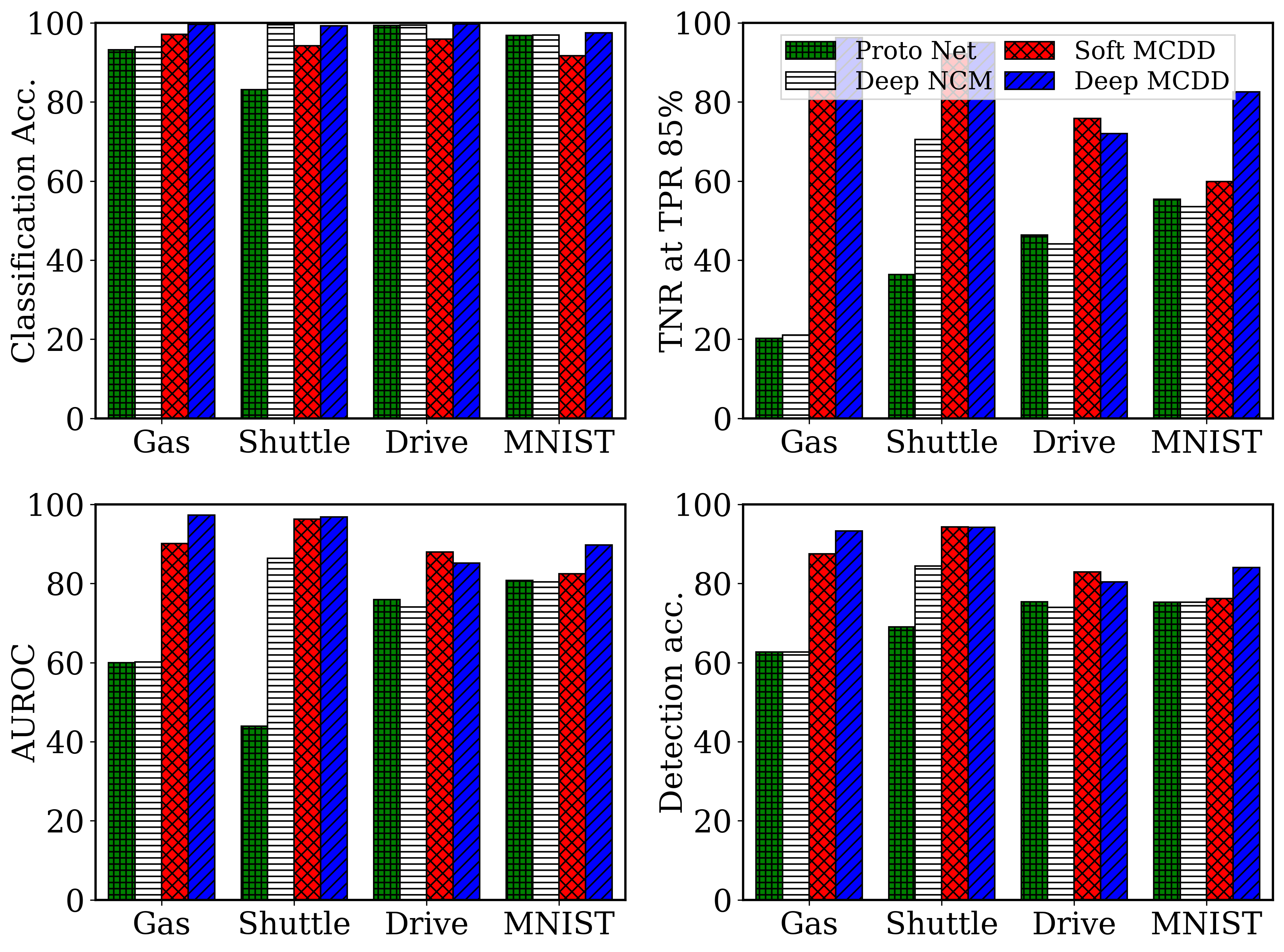}
        \caption{Comparison with distance classifiers which use the Euclidean distance from the nearest class center.}
        \label{fig:compclassifier}
	\end{minipage}
\end{figure*}

In Figure~\ref{fig:compdetector}, our \proposeds show significantly better performances than the baselines in terms of OOD detection, with their stable classification accuracy.   
This implies that class labels are helpful to learn discriminative features between ID and OOD samples as well as among ID samples from different classes.
That is, \proposeds are much more practical for real-world applications compared to non-classifier-based OOD detectors in that it can additionally detect OOD samples while accurately classifying ID samples. 
Moreover, with the help of GDA-based modeling for the multiple spheres, \proposed enables better OOD detection as well as simpler optimization compared to Soft-MCDD.

\subsubsection{Comparison with distance classifiers}
\label{subsubsec:compclassifier}
There have been several attempts to train distance classifiers based on DNNs, so we also consider the following methods as our baselines:
\protonet~\cite{snell2017prototypical} and
\deepncm classifier~\cite{guerriero2018deep}.
Both of them optimize the latent space of DNNs so that the distance of each training sample from its corresponding class center becomes smaller than those from other classes' centers, then predict the class label of an input test sample based on the nearest class center (or class mean).
Their learning objectives are designed only for multi-class classification based on the Euclidean distance, thus they do not take into account OOD samples in the test distribution, as opposed to the baselines considered in Section~\ref{subsubsec:compdetector}.
Since they did not propose confidence scores which can be used for OOD detection, we simply use the Euclidean distance from the nearest class center; i.e., $-\min_k \lVert f(\mathbf{x}) - \spcenter{k} \rVert^2$ where $\spcenter{k}$ is the center of class $k$ used for classification.

In Figure~\ref{fig:compclassifier}, our proposed methods beat the baselines by a large margin for all the metrics.
All the methods classify ID samples correctly to some degree, but only the \proposeds can effectively detect OOD samples while the others fail.
Furthermore, we visualize the 2D latent space obtained by the \deepncm classifier (Figure~\ref{fig:latspace}, Middle).
As shown in the figure, it is obvious that ID and OOD samples in a test distribution considerably overlap with each other.
This degrades the performances by making their confidence scores (i.e., the distances from class centers in the latent space) of ID and OOD samples less apart from each other.

\begin{figure}[t]
    \centering
   	\begin{subfigure}
		\centering
		\includegraphics[width=0.46\linewidth]{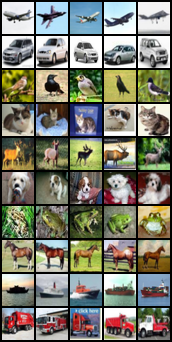}
	\end{subfigure}
	\begin{subfigure}
	    \centering
		\includegraphics[width=0.46\linewidth]{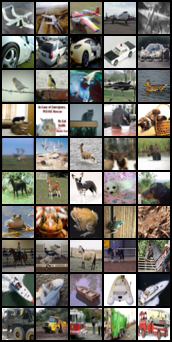}
	\end{subfigure} 
\caption{Test ID images with the highest confidence scores (Left) and the lowest confidence scores (Right) retrieved by \proposed. Dataset: \cifarten, Model: \resnet.}
\label{fig:samples}
\end{figure}

\begin{table}[htbp]
    \caption{ID classification accuracy on image datasets.}
    \label{tbl:imgaccresults}
    \centering
        \begin{tabular}{Scc}
        \hline
        \multirow{2}{*}{Dataset} & \resnet & \densenet \\\cline{2-3}
        & \multicolumn{2}{c}{Softmax / Mahalanobis / \proposed} \\\hline
        \svhn & 95.81 / 95.76 / \textbf{96.55} & 95.30 / 95.22 / \textbf{96.21} \\
        \cifarten & 93.93 / 93.92 / \textbf{95.02} & 92.87 / 91.66 / \textbf{94.97} \\
        \cifarhundred & 75.59 / 74.78 / \textbf{76.15} & 72.27 / 68.22 / \textbf{74.68} \\
        \hline
        \end{tabular}
\end{table}

\begin{table*}[t]
    \caption{Performance of OOD detection on image datasets. The best results are marked in bold face.}
    \label{tbl:imgresults}
    \centering
        \begin{tabular}{QQQcccc}
        \hline
        \multirow{2}{*}{Model} &\multirow{2}{*}{ID} & \multirow{2}{*}{OOD} & TNR at TPR 85\% & AUROC & AUPR & Detection acc.\\\cline{4-7}
        & & & \multicolumn{4}{c}{Softmax~\cite{hendrycks2016baseline} / Mahalanobis~\cite{lee2018simple} / \proposed}  \\\hline
        \multirow{9}{*}{{\resnet}} & \multirow{3}{*}{\svhn} 
          & \cifarten & 89.86 / 90.65 / \textbf{95.41} & 92.29 / 94.81 / \textbf{95.60} & 94.04 / 97.85 / \textbf{98.39} & 87.44 / 87.87 / \textbf{90.42} \\
        & & ImageNet & 91.93 / 85.86 / \textbf{96.75} & 93.58 / 92.94 / \textbf{96.17} & 97.28 / 97.11 / \textbf{98.58} & 88.51 / 85.58 / \textbf{91.60} \\
        & & \lsun & 90.67 / 85.85 / \textbf{95.28} & 92.74 / 92.79 / \textbf{96.02} & 95.44 / 97.13 / \textbf{98.46} & 87.84 / 85.68 / \textbf{91.02} \\\cline{2-7}
        
        & \multirow{3}{*}{\cifarten} 
          & \svhn & 70.85 / 72.21 / \textbf{73.24} & 88.59 / \textbf{88.65} / 86.95 & 82.04 / \textbf{82.86} / 76.84 & 80.75 / \textbf{82.34} / 81.69 \\
        & & ImageNet & 83.53 / 67.76 / \textbf{91.77} & 91.22 / 86.75 / \textbf{94.48} & 93.00 / 88.57 / \textbf{96.24} & 85.01 / 79.21 / \textbf{88.47} \\
        & & \lsun & 89.94 / 73.85 / \textbf{96.41} & 93.19 / 88.95 / \textbf{95.30} & 94.86 / 91.08 / \textbf{96.67} & 87.78 / 82.23 / \textbf{92.92} \\\cline{2-7}
        
        & \multirow{3}{*}{\cifarhundred} 
          & \svhn & 41.14 / 44.52 / \textbf{55.47} & 78.26 / 80.52 / \textbf{81.77} & 70.62 / 67.85 / \textbf{71.35} & 72.69 / 74.44 / \textbf{75.04} \\
        & & ImageNet & 44.49 / 48.04 / \textbf{57.94} & 78.67 / 76.57 / \textbf{83.48} & 81.85 / 77.55 / \textbf{86.48} & 72.21 / 69.58 / \textbf{75.06} \\
        & & \lsun & 44.37 / 46.35 / \textbf{55.94} & 78.87 / 76.41 / \textbf{83.52} & 82.78 / 78.01 / \textbf{87.16} & 72.64 / 69.77 / \textbf{77.06} \\
        \hline
        
        \multirow{9}{*}{{\densenet}} & \multirow{3}{*}{\svhn} 
          & \cifarten & 90.06 / 88.65 / \textbf{94.11} & 92.98 / 93.67 / \textbf{95.76} & 96.13 / 97.44 / \textbf{98.37} & 88.05 / 86.85 / \textbf{89.87} \\
        & & ImageNet & 94.89 / 86.83 / \textbf{95.76} & 95.88 / 93.33 / \textbf{96.83} & 98.01 / 97.10 / \textbf{98.73} & 91.10 / 86.03 / \textbf{91.21} \\
        & & \lsun & 92.63 / 75.94 / \textbf{94.63} & 94.67 / 88.98 / \textbf{96.34} & 97.23 / 95.28 / \textbf{98.55} & 89.70 / 81.24 / \textbf{90.31} \\\cline{2-7}
        
        & \multirow{3}{*}{\cifarten} 
          & \svhn & \textbf{90.63} / 88.32 / 90.42 & 92.87 / \textbf{94.06} / 93.42 & 87.40 / 82.54 / \textbf{88.52} & 87.52 / 87.09 / \textbf{89.55}  \\
        & & Imagenet & 83.98 / 69.47 / \textbf{84.57} & 90.77 / 83.31 / \textbf{90.91} & 88.42 / 76.62 / \textbf{90.39} & \textbf{88.12} / 77.56 / 84.84 \\
        & & \lsun & 85.33 / 66.24 / \textbf{89.28} & 92.26 / 82.82 / \textbf{92.82} & 91.98 / 78.21 / \textbf{93.51} & 84.06 / 75.83 / \textbf{87.26} \\\cline{2-7}
        
        & \multirow{3}{*}{\cifarhundred} 
          & \svhn & 37.80 / 48.96 / \textbf{50.15} & 75.14 / 68.82 / \textbf{77.79} & 58.20 / 62.72 / \textbf{64.14} & 70.03 / 62.02 / \textbf{71.10} \\
        & & ImageNet & 35.22 / 48.21 / \textbf{57.73} & 62.12 / 68.87 / \textbf{81.20} & 57.53 / 76.15 / \textbf{81.93} & 60.30 / 61.73 / \textbf{73.46} \\
        & & \lsun & 38.71 / 43.62 / \textbf{65.11} & 66.36 / 67.51 / \textbf{85.00} & 62.49 / 75.87 / \textbf{86.32}  & 63.15 / 59.94 / \textbf{77.28} \\
        \hline
        \end{tabular}
\end{table*}

\subsection{Evaluation on Image Datasets}
\label{subsec:imgeval}
\smallsection{Experimental settings}
We validate the effectiveness of \proposed on OOD image detection as well.
Two types of deep convolutional neural networks (CNNs) are utilized: \resnet~\citep{he2016deep} with 100 layers and \densenet~\citep{huang2017densely} with 34 layers.
Specifically, we train \resnet and \densenet for classifying three image datasets: \cifarten, \cifarhundred~\citep{krizhevsky2009learning}, and \svhn~\citep{netzer2011reading}.
Each dataset used for training the models is considered as ID samples, and the others are considered as OOD samples. 
To consider a variety of OOD samples at test time, we measure the performance by additionally using \tinyimgnet (randomly cropped image patches of size 32 $\times$ 32 from the ImageNet dataset)~\citep{deng2009imagenet} and \lsun~\citep{yu2015lsun} as test OOD samples.
All CNNs are trained by using SGD with Nesterov momentum~\citep{duchi2011adaptive}, and we follow the training configuration (e.g., the number of epochs, batch size, learning rate and its scheduling, and momentum) suggested by~\citep{lee2018simple, liang2017enhancing}.
The regularization coefficient $\nu$ is set to 0.1 for \cifarten and \cifarhundred, and 0.01 for \svhn.

\begin{figure*}[t]
	\centering
	\includegraphics[width=\linewidth]{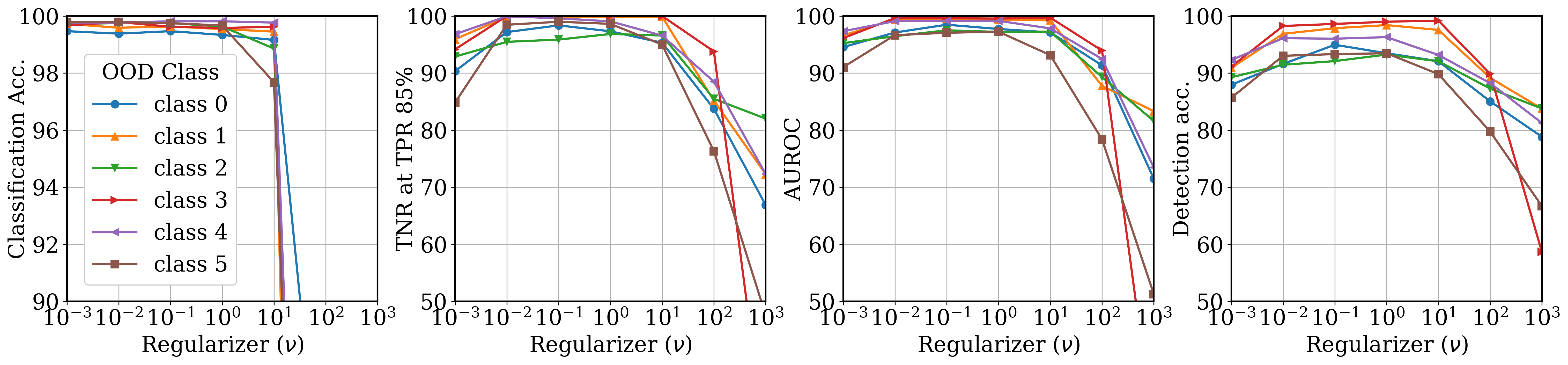}
\caption{The performance changes with respect to $\nu$ values. Dataset: \gas.}
\label{fig:lambdaperform}
\end{figure*}

\smallsection{Experimental results}
Table~\ref{tbl:imgresults} shows that \proposed also can be generalized well for deeper and more complicated DNNs such as \resnet and \densenet.
Similarly to tabular data, our proposed confidence score achieves the best performance for most test cases, and significantly improves the detection performance over the state-of-the-art method.
Interestingly, \proposed achieves slightly better ID classification accuracy than the softmax classifier in Table~\ref{tbl:imgaccresults}.
These results show that any existing DNNs can improve their classification power by adopting the dm-layer, which learns the class means and standard deviations instead of the class weights.
In conclusion, being combined with domain-specific DNNs, our proposed objective enhances the classifier to make reliable predictions for high-dimensional inputs even in case that test distributions differ from the training distribution.

In addition, we provide qualitative results of our proposed confidence score.
Among the test samples from the \cifarten dataset, we retrieve images with the highest and lowest confidence scores (i.e., the samples most likely to come from ID and OOD, respectively) for each class.
In Figure~\ref{fig:samples}, each row contains the retrieved top-5 images for each class.\footnote{The classes are airplane, automobile, bird, cat, deer, dog, frog, horse, ship, and truck.}
The images with high scores share similar global features which can explain each class, whereas the images with low scores look quite different from the prototypical image of each class. 
From the results, we conclude that \proposed is able to not only learn discriminative features among classes but also capture the representative features of the class.

\begin{figure*}[t]
	\centering
	\begin{subfigure}
		\centering
		\includegraphics[width=0.24\linewidth]{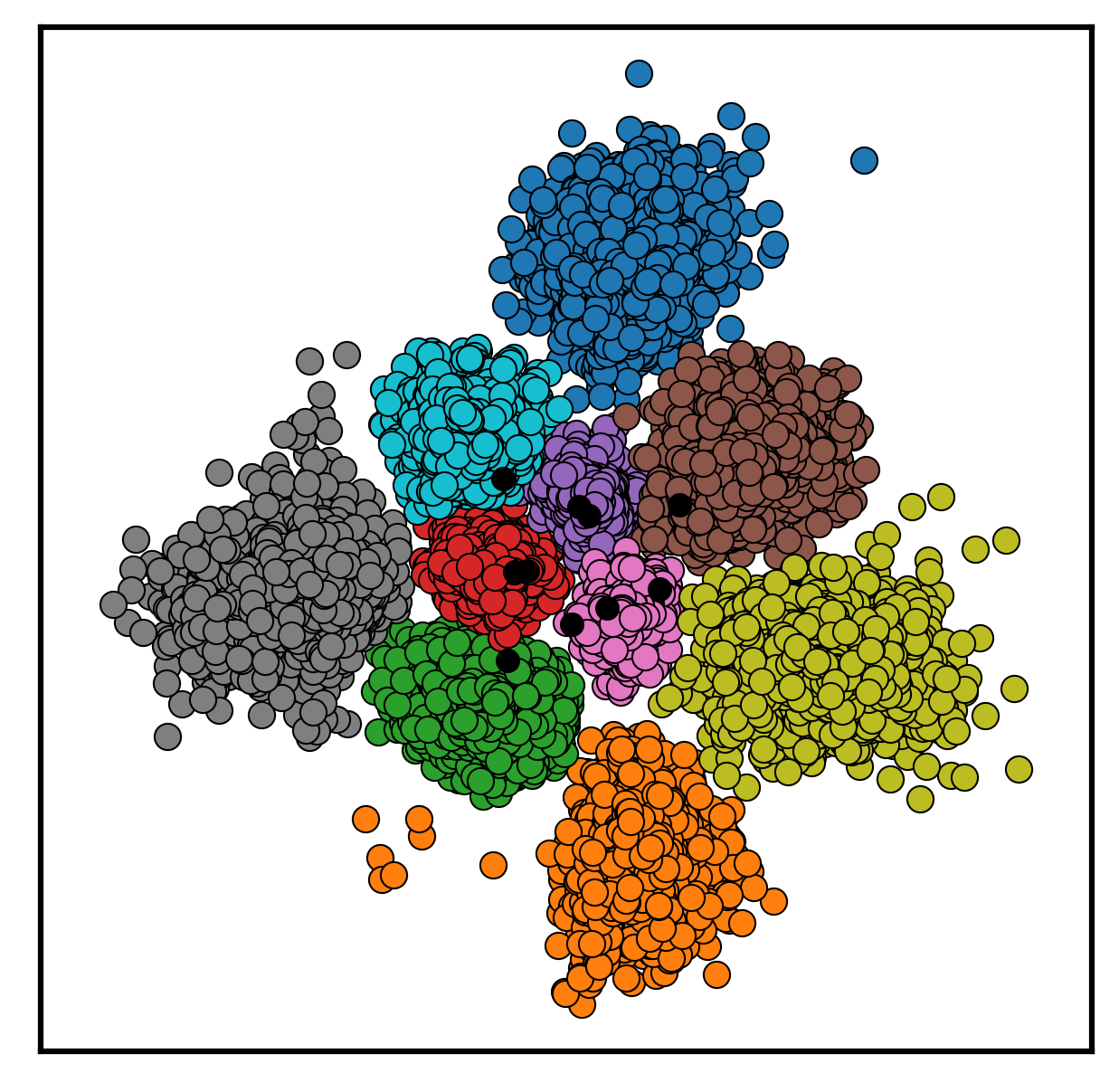}
	\end{subfigure}
	\begin{subfigure}
		\centering
		\includegraphics[width=0.24\linewidth]{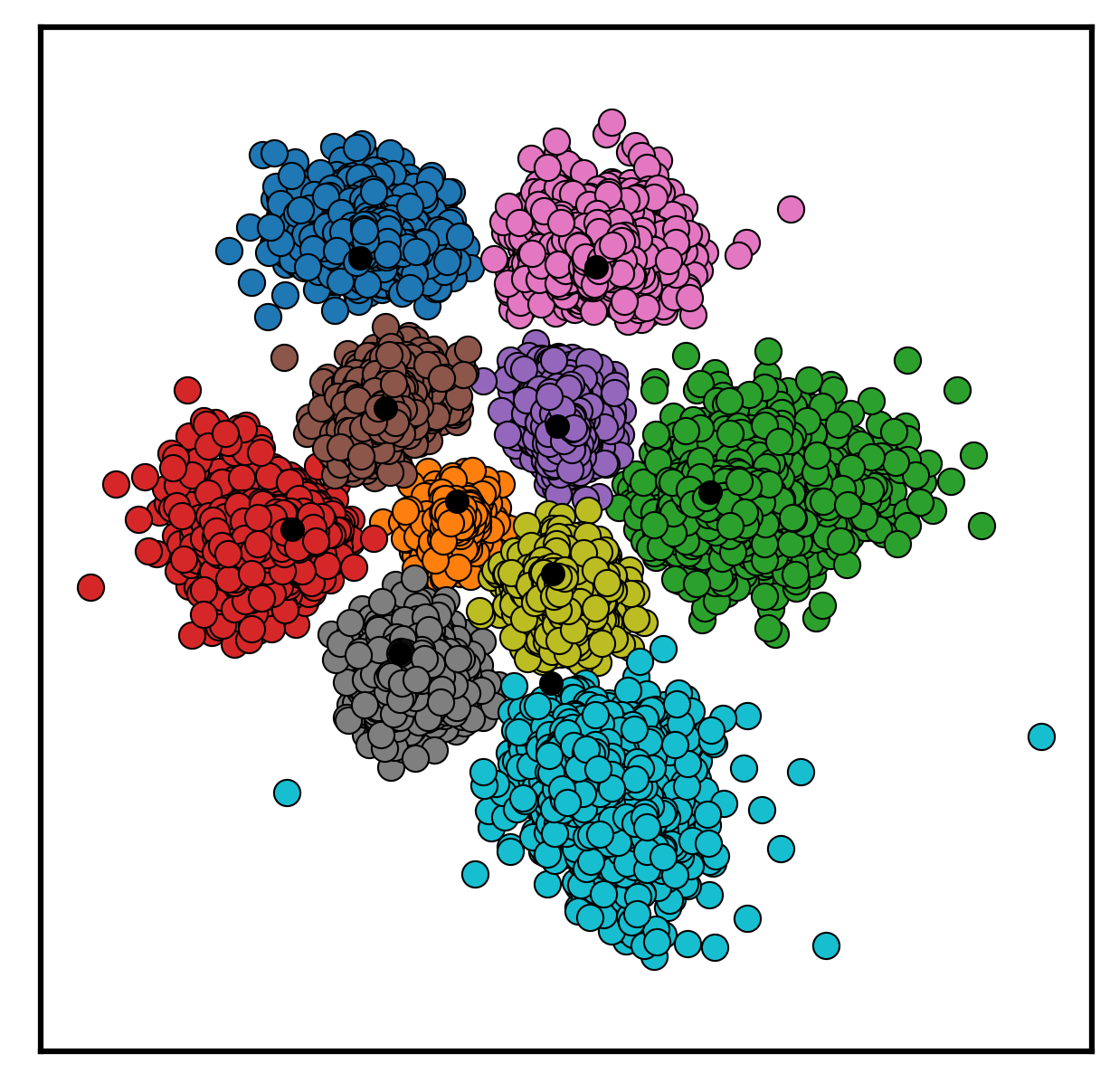}
	\end{subfigure}
	\begin{subfigure}
		\centering
		\includegraphics[width=0.24\linewidth]{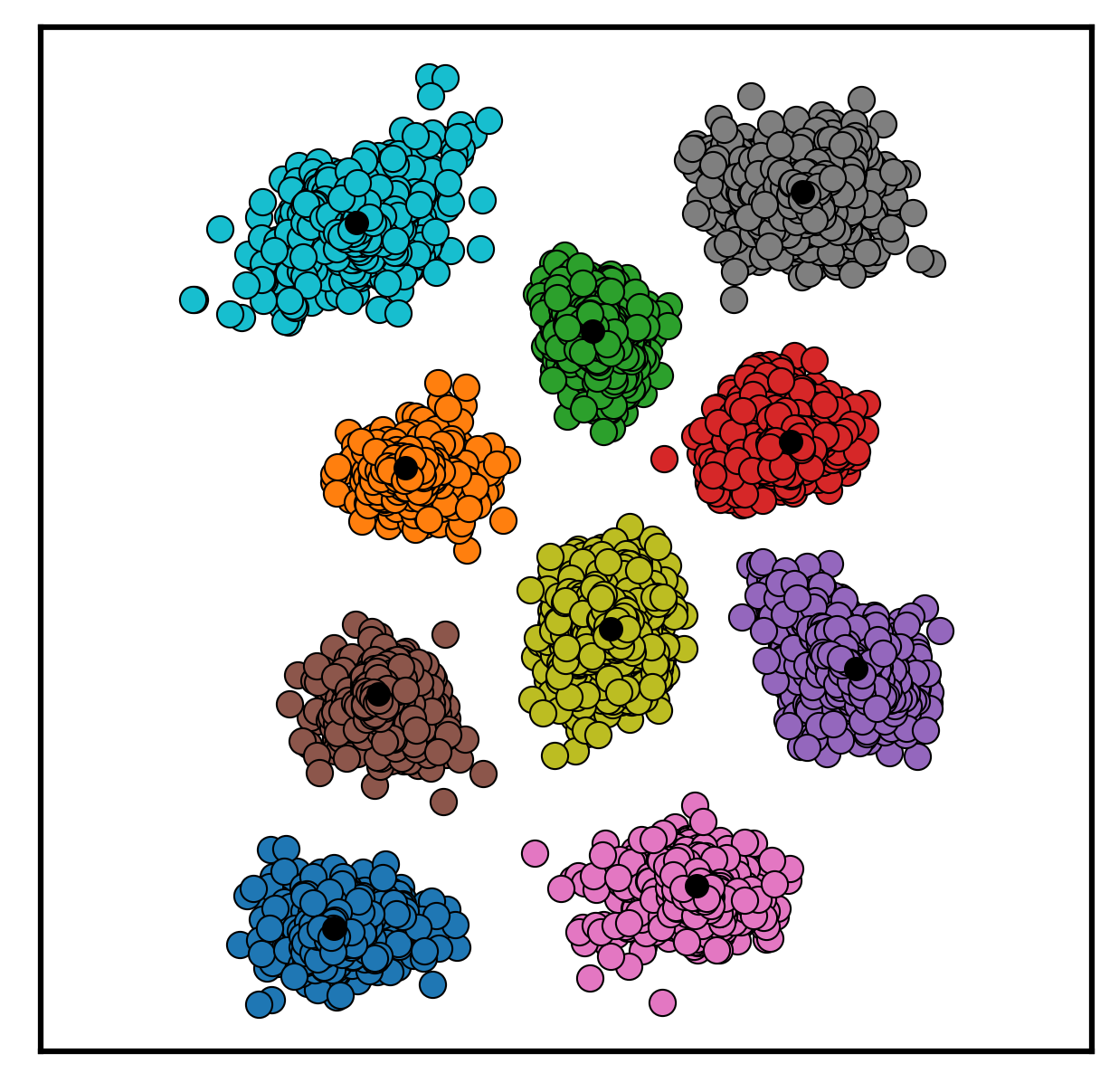}
	\end{subfigure}
	\begin{subfigure}
		\centering
		\includegraphics[width=0.24\linewidth]{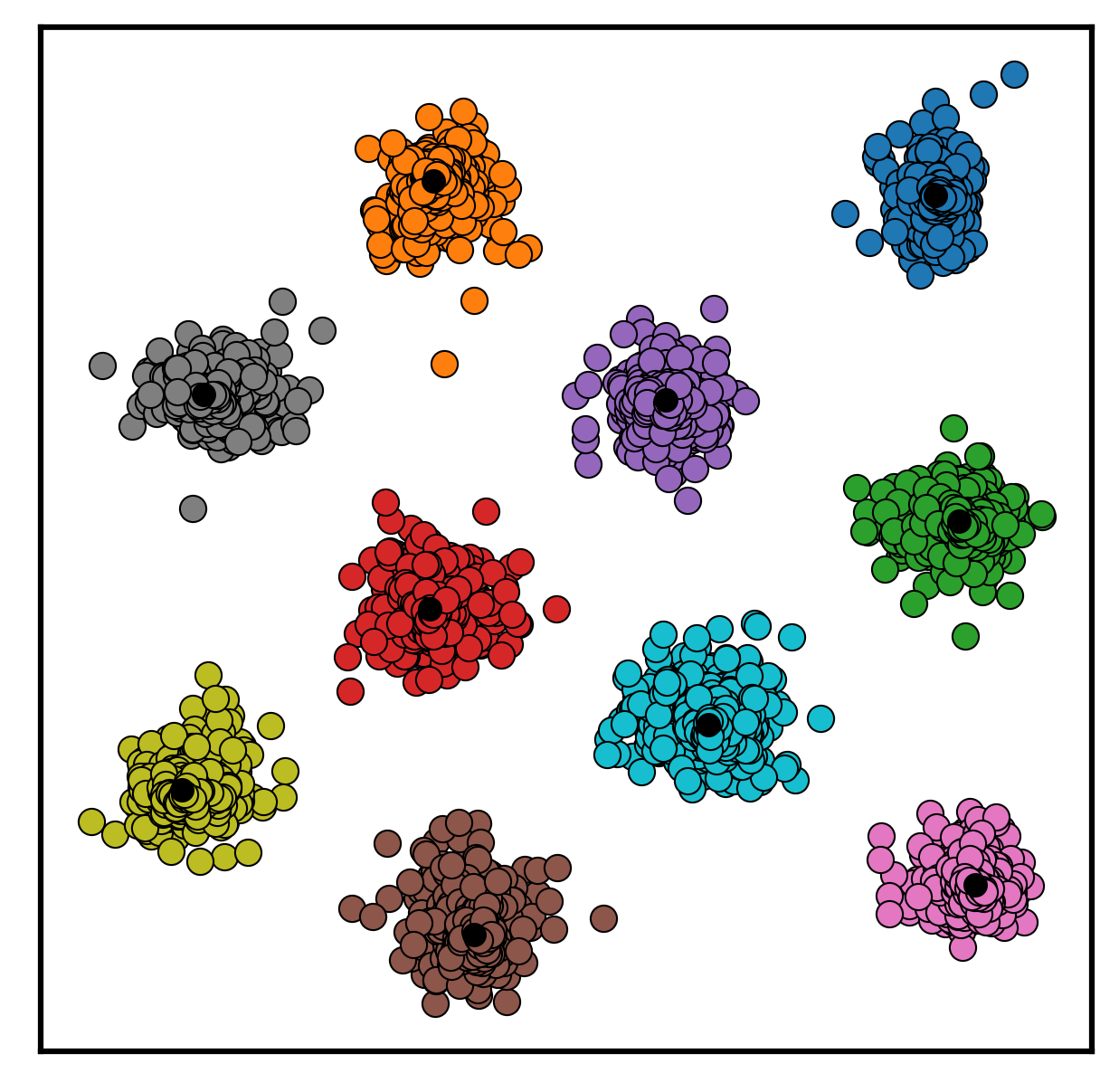}
	\end{subfigure}
\caption{The training distribution in our 2D latent space, $\nu=$ 0.001, 0.01, 0.1, and 1. Dataset: \mnist.}
\label{fig:lambdaspace}
\end{figure*}

\subsection{Effect of Regularization}
\label{subsec:regeffect}

We further investigate the effects of our regularization term on the performance and the data distributions in the latent space.
We first evaluate \proposed with different $\nu$ values from $10^{-3}$ to $10^3$.
Figure~\ref{fig:lambdaperform} presents the performance changes with respect to the $\nu$ value.
In terms of ID classification, \proposed cannot be trained properly when $\nu$ grows beyond $10^2$, because the KL divergence term is weighted too much compared to the log posterior term which learns the decision boundary for multi-class classification.
On the other hand, we observe that the OOD detection performances are not much affected by the regularization coefficient, unless we set $\nu$ too small or too large; any values in the range (0.01, 10) are fine enough to obtain the model working well.

We also visualize the 2D latent space where the training distribution of \mnist are represented, varying the value of $\nu \in\{0.001, 0.01, 0.1, 1\}$.
In Figure~\ref{fig:lambdaspace}, even with a small value of $\nu$, we can find the decision boundary that partitions the space into $K$ regions, whereas the class means (plotted as black circles) do not match with the actual class means and the samples are spread over a broad region in the latent space.
As $\nu$ increases, the class means approach to the actual class means, and simultaneously the samples get closer to its corresponding class mean thereby form multiple spheres.
As discussed in Section~\ref{subsec:obj}, the coefficient $\nu$ controls the trade-off between separating each class-conditional distribution from the others and enforcing them to approximate the Gaussian distribution with the mean $\mean{k}$.
In summary, the proper value of $\nu$ makes the DNNs place the class-conditional Gaussian distributions far apart from each other, so the OOD samples are more likely to be located in the rest of the space.

\section{Conclusion}
\label{sec:conc}
This paper introduces deep learning objectives for multi-class data description, which aims to find the spherical decision boundary for each class in a latent space.
By fusing the concept of Gaussian discriminant analysis with DNNs, \proposed learns the class-conditional distributions to be separated from each other and follow the Gaussian distribution at the same time, thus it is able to effectively identify OOD samples that do not belong to any known classes.
We empirically show that \proposed outperforms competing methods in detecting both OOD tabular data and OOD images, and also the proposed dm-layer can be easily combined on top of various types of DNNs to further improve their performances.

\begin{acks}
This work was supported by the NRF grant funded by the MSIT: (No.~2016R1E1A1A01942642) and (No.~2017M3C4A7063570), the IITP grant funded by the MSIT: (No.~2018-0-00584), (No.~2019-2011-1-00783), and (No.~2019-0-01906).
\end{acks}

\bibliographystyle{ACM-Reference-Format}
\bibliography{bibliography}

\end{document}